\documentclass{article}
\usepackage[margin=1.2in]{geometry}

\usepackage[utf8]{inputenc} 
\usepackage[T1]{fontenc}    
\usepackage{hyperref}       
\usepackage{url}            
\usepackage{booktabs}       
\usepackage{amsfonts}       
\usepackage{amsmath}
\usepackage{nicefrac}       
\usepackage{microtype}      
\usepackage{graphicx}
\usepackage{caption}
\usepackage{subcaption}
\usepackage{natbib}
\usepackage{color}

\newcommand{\githubenn}{\url{https://github.com/deepmind/enn}}

\newcount\Comments
\newcommand{\kibitz}[2]{\ifnum\Comments=1{\textcolor{#1}{\textsf{\footnotesize #2}}}\fi}

\DeclareMathOperator*{\softmax}{softmax}

\title{Robustness of Epinets against Distributional Shifts \vspace{0.3cm}}

\author{
  \bf \normalsize
  Xiuyuan Lu\thanks{Contact \texttt{lxlu@deepmind.com}.},
  Ian Osband,
  Seyed Mohammad Asghari,
  Sven Gowal,\\
  \bf \normalsize
  Vikranth Dwaracherla,
  Zheng Wen,
  Benjamin Van Roy \\[0.3cm]
  \normalsize
  DeepMind
}

\date{\vspace{-0.5cm}}

\begin{document}

\maketitle

\begin{abstract}
Recent work introduced the \textit{epinet} as a new approach to uncertainty modeling in deep learning \citep{osband2022epistemic}.
An epinet is a small neural network added to traditional neural networks, which, together, can produce predictive distributions.
In particular, using an epinet can greatly improve the quality of \textit{joint} predictions across multiple inputs, a measure of how well a neural network knows what it does not know \citep{osband2022epistemic}.
In this paper, we examine whether epinets can offer similar advantages under distributional shifts.
We find that, across ImageNet-A/O/C, epinets generally improve robustness metrics. Moreover, these improvements are more significant than those afforded by even very large ensembles at orders of magnitude lower computational costs.
However, these improvements are relatively small compared to the outstanding issues in distributionally-robust deep learning.
Epinets may be a useful tool in the toolbox, but they are far from the complete solution.
\end{abstract}

\section{Introduction}
Epistemic neural networks (ENNs) were recently introduced as a new framework for modeling uncertainty in deep learning \citep{osband2022epistemic}. ENNs offer an interface for expressing uncertainty due to knowledge, the kind that can be resolved with additional data, as opposed to uncertainty due to chance. The ability to express uncertainty due to knowledge is crucial to intelligence. For example, effective exploration, adaptation, and decision making should rely on an agent knowing what it does not know.

Under the ENN framework, the paper introduces a new architecture, called the {\it epinet}, for uncertainty modeling \citep{osband2022epistemic}. An epinet is a relatively small neural network added to a big ``base network'' to produce uncertainty estimates. The base network can have any modern deep learning architecture, and it can even be a pre-trained network. The epinet is designed to be economic in the amount of computation it requires (typically far less than the computation required by the base network), while delivering performance on par or better than popular Bayesian deep learning approaches. That paper shows that epinets perform well on a range of image classification and reinforcement learning tasks, either in the statistical quality or computational requirements, or both, compared to alternative approaches. In particular, the paper shows that neural networks with an epinet can outperform very large ensembles at orders of magnitude lower computational costs.

Following these promising results, one natural question to ask such a network, which is designed to know what it doesn't know, is:
\begin{center}
{\it Does an epinet offer any statistical or computational benefits in tasks with distributional shifts?}
\end{center}
While it is possible to design versions of epinets specifically to address distributional shifts, we defer those investigations to future research.
As a first step, we take the epinet trained on ImageNet by \cite{osband2022epistemic} and study its robustness on a set of ImageNet distributional-shift benchmarks, including ImageNet-A, ImageNet-O, and ImageNet-C \citep{hendrycks2021nae, hendrycks2019robustness}. The epinet is trained on top of a base network that is a pre-trained ResNet. We compare the epinet against its base ResNet, as well as an ensemble of ResNets, a popular approach to uncertainty estimation. 
More specifically, we are interested in knowing whether the epinet improves the robustness of its associated base network on these benchmarks, and whether the epinet is statistically or computationally more efficient than the ensemble approach for handling distributional shifts.

In addition to traditional measures of robustness, we also measure the quality of {\it joint predictions} across multiple inputs.
A key result in \cite{osband2022epistemic} is that the epinet is able to provide joint predictions of much higher quality than alternative approaches. The paper points out that the quality of joint predictions is a measure of how well the neural network knows that it does not know.
For example, in the context of out-of-distribution inputs, consider an agent that is uncertain about the labels associates with these inputs. By looking at the agent's predictive distribution at a single input, one cannot tell whether the agent's uncertainty would resolve if trained on this data point. However, as pointed out in \cite{osband2022epistemic}, by looking at joint predictions across multiple inputs, one can distinguish whether the agent's uncertainty would resolve if trained on these out-of-distribution examples.
This knowledge can be particularly useful if the agent plans to gather more data to improve its predictions.
Further, as elaborated in \cite{wen2022predictions}, the quality of joint predictions has important relevance to decision making.
Thus, a particular interesting question to look at is whether the epinet makes better joint predictions than alternative approaches, not only on in-distribution test data like what we see in \cite{osband2022epistemic}, but also on test data with distributional shifts.

Here is a summary of our key observations on ImageNet-A/O/C.
\begin{enumerate}
\item The epinet improves or performs similarly to the associated ResNet according to traditional robustness metrics. However, it does not completely address these distributionally-robust challenges. This is not surprising, as the current epinet architecture and training are not designed to address such challenges.

\item The ensemble approach is not competitive with the epinet in either the statistical qualities or computational costs according to traditional robustness metrics.

\item The epinet dramatically outperforms the ResNet and ensemble baselines in joint predictions, at a computational cost not much more than that of the base ResNet. This is similar to the results for in-distribution test data in \cite{osband2022epistemic}.
\end{enumerate}

Our results point to an important future research direction. Even though the epinet is more robust against distributional shifts compared to the ResNet and ensemble baselines, as it is currently trained, it does not completely address these challenges. We believe that a stronger prior is needed to inform the agent of the possibility of out-of-distribution inputs. This can, for example, take form in various types of data augmentation, which is a common heuristic in the literature. It is also worthwhile to design and investigate other forms of regularization.
Fortunately, the ENN framework can easily accommodate these techniques, while offering the additional benefit of allowing neural networks to express knowledge of what they do and do not know.

\section{Experimental setup}
We describe the experimental setup for evaluation on ImageNet-A/O/C. We will discuss the models, datasets, and evaluation metrics.

\subsection{Models}
We evaluate the ResNet, epinet, and ensemble models from \cite{osband2022epistemic}. The models are trained on the ImageNet dataset \citep{deng2009imagenet}. The ResNet and epinet architectures, together with their checkpoints, are available in the open-source library at \githubenn.

For the ResNet model, we consider several standard architectures, including ResNet-$L$ for $L \in \{50, 101, 152, 200\}$. We consider ensembles made from ResNet-$50$ models as ensemble members. We independently train $30$ ResNet-$50$ models and use them to form ensembles of size $1$, $3$, $10$, and $30$.

Recall from \cite{osband2022epistemic} that an ENN can be described by a tuple $\left(f_\theta(x, z), P_Z\right)$, where $f$ is a network with parameters $\theta$ that takes input $x$ and an \textit{epistemic index} $z$, and $P_Z$ is a reference distribution from which the epistemic index $z$ is drawn.
For a single input $x$, an ENN would assign a probability $\int_z P_Z(dz) \softmax \left(f_{\theta}(x, z)\right)_{y}$ to each class $y$. To make a joint prediction across multiple inputs $(x_1, \dots, x_\tau)$, an ENN would assign a probability
\[ \int_z P_Z(dz) \prod_{t=1}^\tau \softmax \left(f_{\theta}(x_t,z)\right)_{y_t} \]
to each class combination $(y_1, \dots, y_\tau)$. Note that by introducing dependencies on the epistemic index $z$, ENNs allow for more expressive joint predictions beyond simply the product of marginal predictions.

Under the ENN framework, the epinet approach can be described as
\begin{equation}
\label{eq:epinet}
f_{\zeta, \eta}(x, z)
= \underbrace{\mu_\zeta(x)}_{\textrm{\footnotesize base net}} + \underbrace{\sigma_\eta(\phi_\zeta(x), z)}_{\textrm{\footnotesize epinet}}
= \underbrace{\mu_\zeta(x)}_{\textrm{\footnotesize base net}} + \underbrace{\sigma_\eta^L(\phi_\zeta(x), z)}_{\textrm{\footnotesize learnable epinet}} + \underbrace{\sigma^P(\phi_\zeta(x), z)}_{\textrm{\footnotesize prior epinet}},
\end{equation}
where $z$ is the epistemic index with a standard Gaussian reference distribution, the base net $\mu_\zeta$ is a pre-trained ResNet with parameters $\zeta$, and $\phi_\zeta(x)$ denotes information from the base net that is passed as input to the epinet. Specifically, $\phi_\zeta(x)$ includes the input image $x$ as well as the last-layer features from the base ResNet. 
The epinet is composed of a learnable network $\sigma_\eta^L$ with weights $\eta$ and a fixed prior network $\sigma^P$.
The learnable network is a variant of an MLP, and the prior network is a combination of an MLP variant and an ensemble of small convolutional networks.
The sizes of the learnable epinet and prior epinet are several orders of magnitude smaller than the size of the base ResNet.
An epinet is trained for each ResNet-$L$ base network for $L \in \{50, 101, 152, 200\}$.
The network $f_{\zeta, \eta}$ outputs logits and is trained using cross-entropy loss and ridge regression. The weights $\zeta$ of the base network are frozen during training.
More details about the architecture and training can be found in Section 3.2 and Appendix F.1 of \cite{osband2022epistemic}, as well as in the open-source library.
To obtain a predictive distribution from the network $f_{\zeta, \eta}$, we sample $1000$ epistemic indices from the reference distribution and average over the corresponding predictions.

\subsection{Datasets}
We consider several standard ImageNet robustness benchmarks, including ImageNet-A and ImageNet-O from \cite{hendrycks2021nae}, as well as ImageNet-C from \cite{hendrycks2019robustness}. 
ImageNet-A concerns a collection of adversarial images that are selected to trick a ResNet-50 ImageNet classifier to make the wrong prediction.
ImageNet-O is a collection of images that do not belong to the $1000$ classes that appear in the ImageNet training set.
ImageNet-C concerns a collection of synthetic corruptions that are applied to the ImageNet test images. It includes $16$ types of corruption noise and $5$ levels of corruption severity.

For ImageNet-A and ImageNet-O, subsets with size $200$ from the $1000$ ImageNet classes are selected so that a misclassification would be considered egregious \citep{hendrycks2021nae}. Following \cite{hendrycks2021nae}, we compare models that make predictions on these subsets of classes. To restrict our benchmark models to predict a subset of classes, we restrict the logits to these classes before taking the softmax.

\subsection{Metrics}
For ImageNet-A, we evaluate the prediction accuracy, expected calibration error, and two types of log-losses, the marginal log-loss and joint log-loss. The marginal log-loss is the expected negative log-likelihood of a \textit{single} test example under the model's predictive distribution for the single test input. The joint log-loss is the expected negative log-likelihood of a \textit{batch} of test examples under the model's \textit{joint} predictive distribution, which is over combinations of labels, for the whole batch of inputs. We take the batch size to be $10$, and we apply the dyadic sampling heuristic from \cite{osband2022evaluating} to measure the joint log-loss.

For ImageNet-O, the goal is for the model to distinguish out-of-distribution test images from the in-distribution test images taken from the ImageNet test set (restricted to the $200$ classes mentioned above). We follow \cite{hendrycks2021nae} and measure the area under the precision-recall curve (AUPR). For each image, the anomaly score is defined as the negative of the maximum class probability. The anomaly scores for these in-distribution and out-of-distribution test images are used to compute the AUPR.

For ImageNet-C, the labels of the corrupted images are taken to be the same as the original images. We measure the prediction error, expected calibration error, marginal log-loss, and joint log-loss for each combination of corruption type and severity. Following \cite{hendrycks2019robustness}, for the prediction error, we first sum the prediction errors over corruption severities for each corruption type, and then take a weighted average of the sums over all the corruption types, where the weights are the inverse of the summed prediction errors obtained by an AlexNet. The weighted average is referred to as the mean corruption error, or mCE, in \cite{hendrycks2019robustness}. For the calibration error, marginal, and joint log-losses, we take a simple average over all the corrupted datasets.

Since we evaluate the model checkpoints and do not re-train any models, the only source of uncertainty in evaluation comes from the random sampling of epistemic indices. We have found that the randomness has a negligible effect on our results, and so, to simply the figures, we will omit the error bars. On the other hand, how the randomness in model training affects these metrics is left for future research.

\section{Results}
We present our main results for ImageNet-A, ImageNet-O, and ImageNet-C in Sections~\ref{se:imagenet-a}, \ref{se:imagenet-o}, and \ref{se:imagenet-c}, respectively. These results demonstrate that, on ImageNet-A/O/C, the epinet improves or is around the same level of robustness compared to the associated ResNet baseline according to traditional evaluation metrics involving only marginal predictive distributions. 
The ensemble baseline, interestingly, is not competitive with the epinet according to these metrics as we increase the model size for both methods.
For joint predictions, the epinet significantly outperforms the ResNet and ensemble baselines. 
That said, we will see that even though using an epinet helps with robustness in general, it does not fully address the challenges presented by these datasets, which deserve future research. 

We follow up our investigation of model behaviors in Section~\ref{se:prediction-uncertainty}, where we look at the prediction confidence of these benchmark models on ImageNet-A/O/C. We will see that these models, in fact, demonstrate reasonable levels of prediction confidence on ImageNet-A and ImageNet-C, but all of them are over-confident on ImageNet-O.

A widely used heuristic to improve model performance on the ImageNet test set is to re-scale the logits using a tunable temperature parameter post training. \citet{osband2022epistemic} also finds that this heuristic improves model performance on ImageNet. In Section~\ref{se:temperature}, we will look at how applying these temperatures tuned for ImageNet evaluation affects model robustness on ImageNet-A/O/C. We will see that re-scaling model predictions by these temperatures in general does not improve robustness, and, in a few cases, even hurts model robustness by a significant amount.

\subsection{ImageNet-A}
\label{se:imagenet-a}
\begin{figure}[b]
\centering
\includegraphics[width=\textwidth]{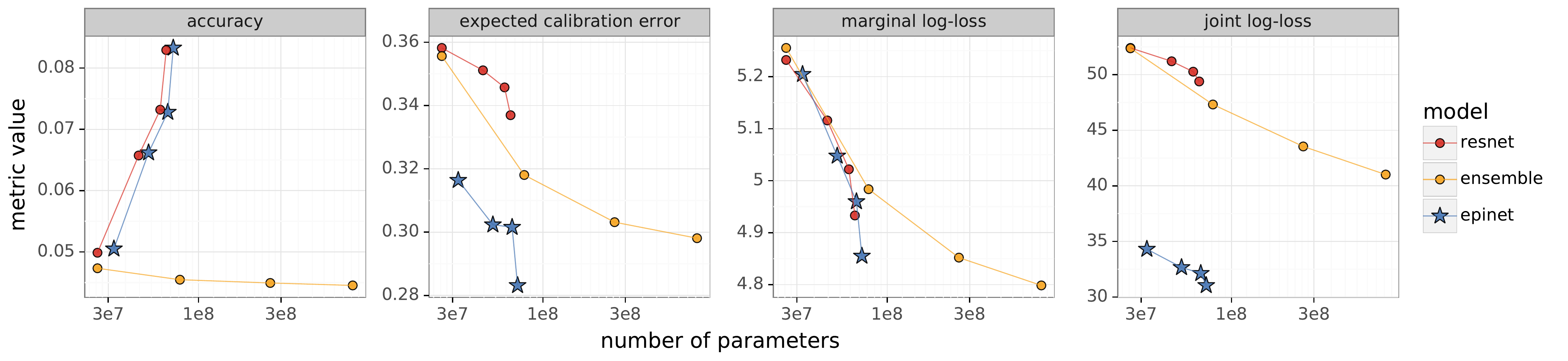}
\caption{Model performance on ImageNet-A.}
\label{fig:imagenet_a}
\end{figure}
Figure~\ref{fig:imagenet_a} shows the performance of the ResNet, epinet, and ensemble models on ImageNet-A.
We show the accuracy, calibration error, marginal log-loss, and joint log-loss of these models as we increase the model size.
In each plot, the four ResNet data points correspond to ResNet-$L$ for $L \in \{50, 101, 152, 200\}$. The four epinet data points correspond to the four epinets trained on these ResNet base networks, respectively. The four ensemble data points correspond to ensembles of sizes $1$, $3$, $10$, and $30$.
We observe that bigger networks in general improve model performance, with the exception that the accuracy of ensembles does not increase with the ensemble size. This could be due to the fact that ImageNet-A is designed to trick a ResNet-$50$ model, and an ensemble of ResNet-$50$ models might still fail to improve accuracy.

We see that the epinet achieves similar accuracy as its associated base ResNet. Both models significantly outperform the ensemble in accuracy as the model size increases.
The epinet improves the calibration error over the base ResNet. Even though ensembling also improves the calibration error, the improvement is not as much for a fixed model size compared to the epinet.
The epinet slightly improves the marginal log-loss over the base ResNet, but a similar improvement can be achieved by increasing the ResNet size. For a fixed model size, the ensemble does not seem to offer any benefit in the marginal log-loss over the other approaches.
For the joint log-loss, the epinet outperforms alternatives by a significant margin. This huge gap in performance echoes what \citet{osband2022epistemic} observe with in-distribution evaluation.

Overall, the epinet improves the robustness over the base ResNet while requiring only a little additional computation, and it offers a huge advantage in the quality of joint predictions compared to the baseline approaches. However, we see that the accuracy of the epinet model is below $10\%$ -- it is still far from solving the challenge of correctly classifying these adversarial images. That said, this particular epinet model is not designed to address such a challenge. We hypothesize that the epinet (or other ENNs), together with its base net, need to be equipped with a stronger prior, either in training or architecture, in order to address this issue, which we defer to future research.

\subsection{ImageNet-O}
\label{se:imagenet-o}
\begin{figure}
\centering
\includegraphics[width=0.4\textwidth]{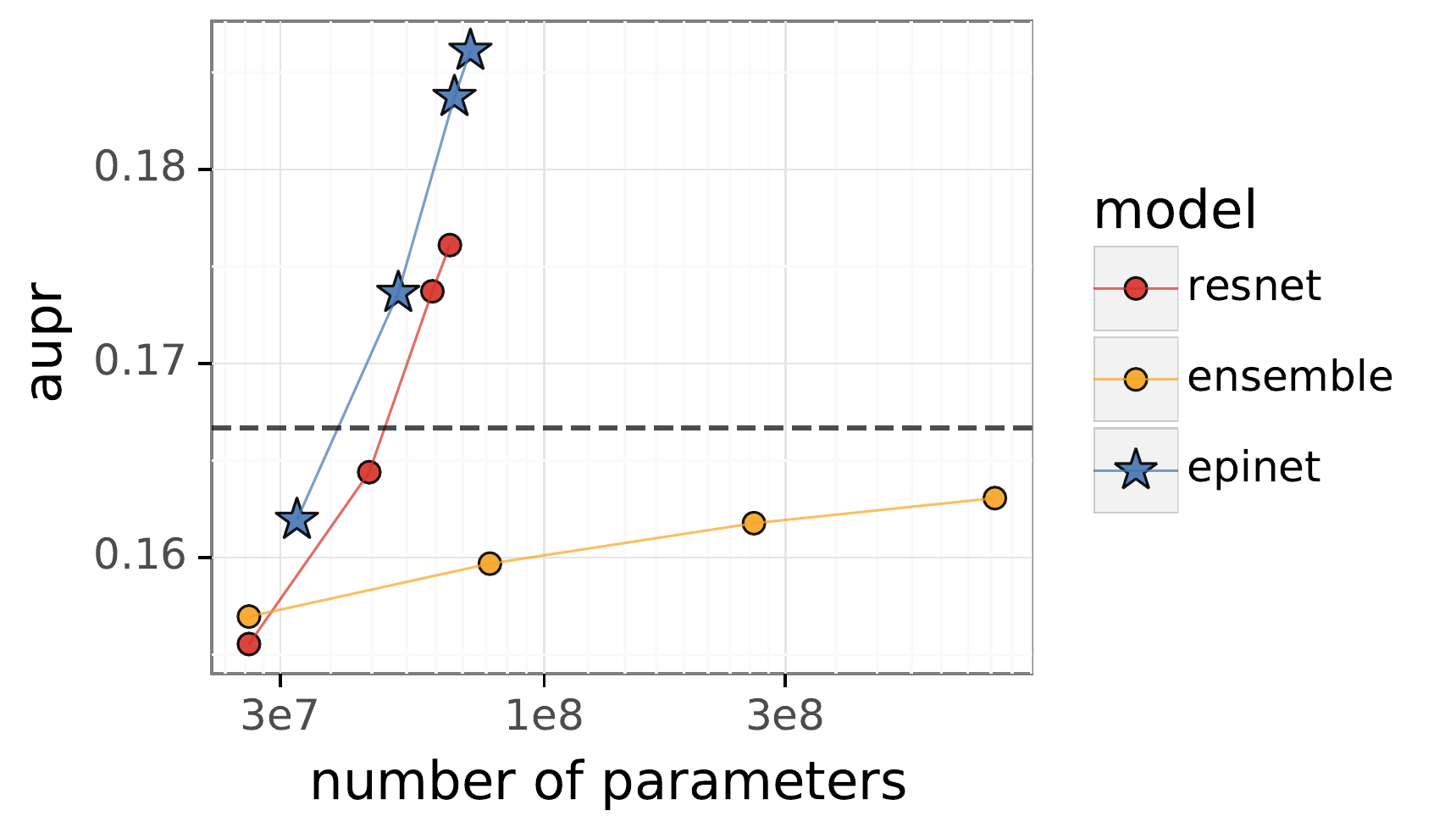}
\caption{Model performance on ImageNet-O. The dashed line corresponds to a uniformly random classifier.}
\label{fig:imagenet_o}
\end{figure}
Figure~\ref{fig:imagenet_o} presents the results on ImageNet-O. We show the area under the precision-recall curve of the benchmark models as their model sizes increase. We see that all models perform better with a larger model size. The epinet outperforms the ResNet, both of which perform better than the ensemble. Again, we see that even though the epinet improves the robustness of the base ResNet, it is not much better than a uniformly random classifier, shown as the dashed line in Figure~\ref{fig:imagenet_o}. Clearly, additional techniques are needed to fully address the challenge imposed by this dataset.

\subsection{ImageNet-C}
\label{se:imagenet-c}
\begin{figure}
\centering
\includegraphics[width=\textwidth]{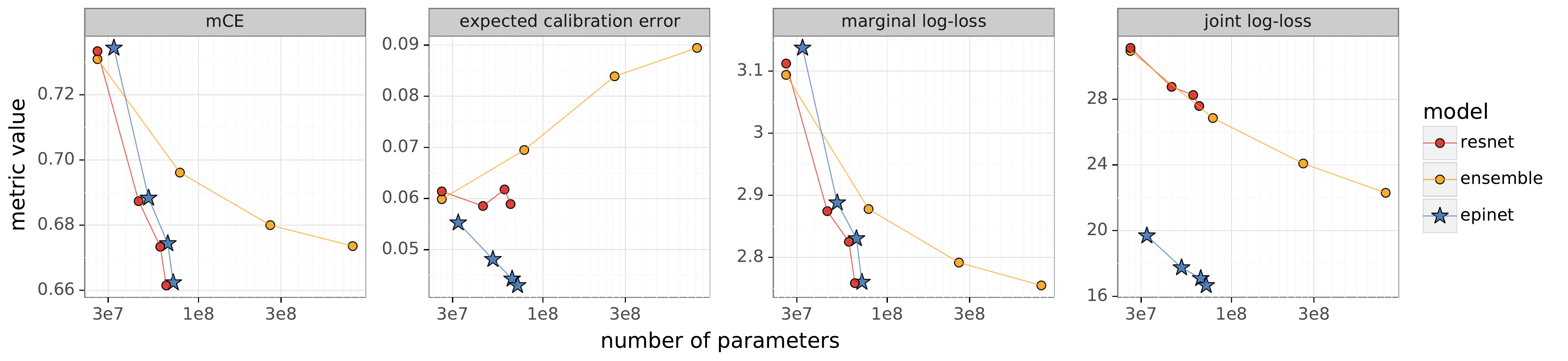}
\caption{Model performance on ImageNet-C averaged over tasks. The mean weighted corruption error (mCE) is the average test error weighted by the errors of an AlexNet. For the other metrics, the values are a simple average across all tasks.}
\label{fig:imagenet_c}
\end{figure}
Figure~\ref{fig:imagenet_c} presents the main results on the ImageNet-C dataset.
The figure shows the mean weighted corruption error (mCE), expected calibration error, marginal log-loss, and joint log-loss as a function of the model size.
Each data point is averaged over the corruption types and corruption severities.

The epinet performs similarly to the ResNet in mCE. Both methods get better mCE with an increasing model size, and both methods improve the metric more so than the ensemble as the model size increases.
The epinet is slightly better than the ResNet at the expected calibration error.
Interestingly, the calibration error of the ensemble model degrades as the ensemble size increases. This seems similar to the observations made by \citet{rahaman2021ensemble}, who observe a similar phenomenon and point out that this degradation would go away by scaling the logits using a cold temperature. We make an analogous observation in Section~\ref{se:temperature} where we investigate temperature re-scaling.
For the marginal log-loss, the epinet is no better than its base ResNet. Increasing the ensemble size improves the marginal log-loss, but we can obtain a larger improvement by increasing the model size of the ResNet and epinet models.
Similar to ImageNet-A, the epinet dramatically outperforms the ResNet and ensemble baselines in the joint log-loss.

We see here that the epinet attains around the same level of robustness as its base ResNet on ImageNet-C according to mCE, calibration error, and marginal log-loss. Ensembling, interestingly, is not a competitive approach according to these metrics. For joint predictions, epinet demonstrates a significant advantage over the other agents, similar to what we see in ImageNet-A (Section~\ref{se:imagenet-a}) and in-distribution evaluation \citep{osband2022epistemic}.

\subsection{Prediction uncertainty}
\label{se:prediction-uncertainty}

We dive deeper into model behaviors by looking at the uncertainty of model predictions on these datasets. For a specific model, we define the confidence score as the average probability assigned to the predicted labels. Note that the confidence score relies only on a model's marginal predictions.

\begin{figure}
\centering
\begin{subfigure}[t]{0.45\textwidth}
    \centering
    \includegraphics[width=\textwidth]{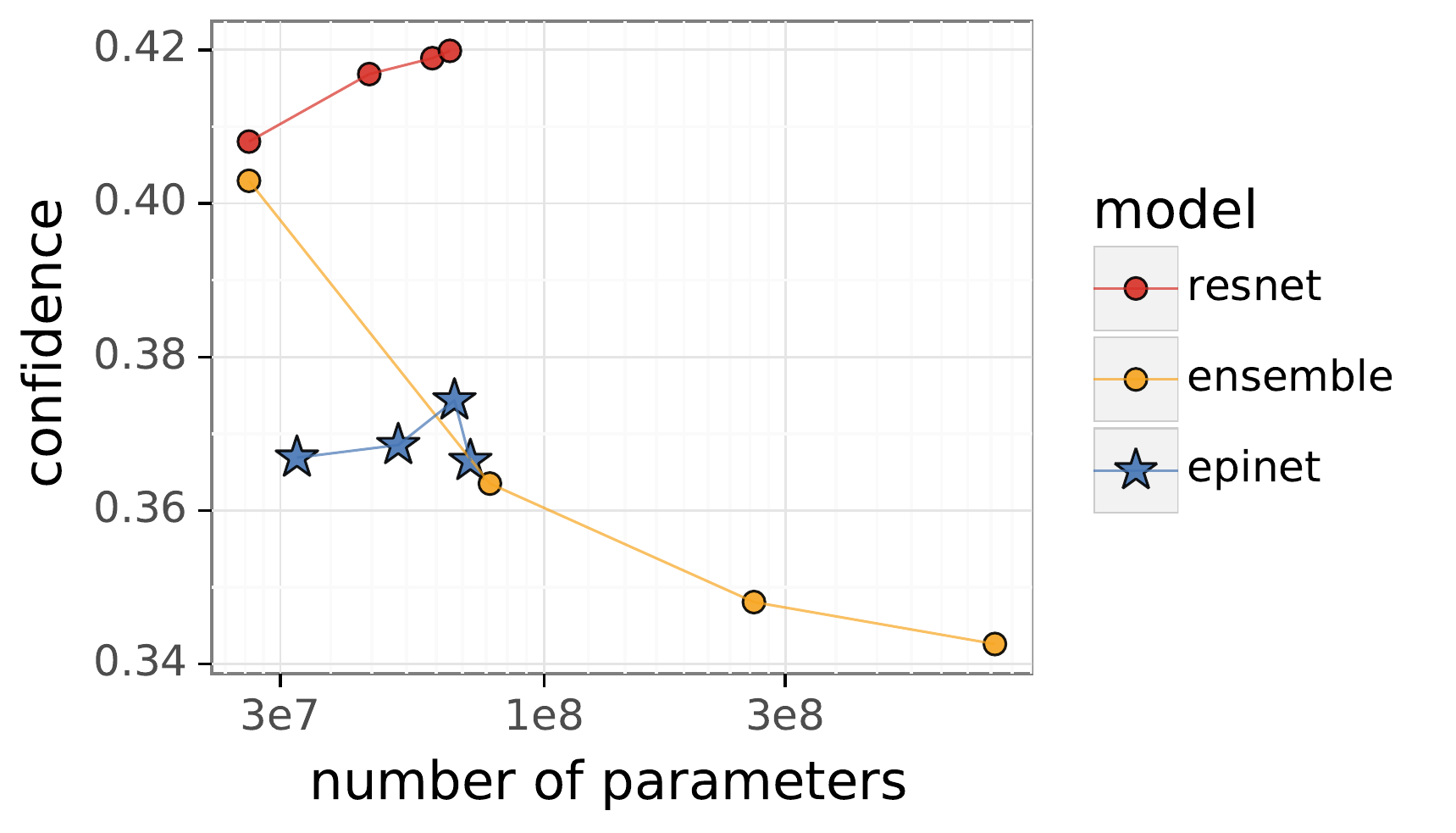}
    \caption{Model prediction confidence, which is the average probability assigned to predicted labels, on ImageNet-A.}
    \label{fig:imagenet_a_confidence}
\end{subfigure}
~
\begin{subfigure}[t]{0.45\textwidth}
    \centering
    \includegraphics[width=\textwidth]{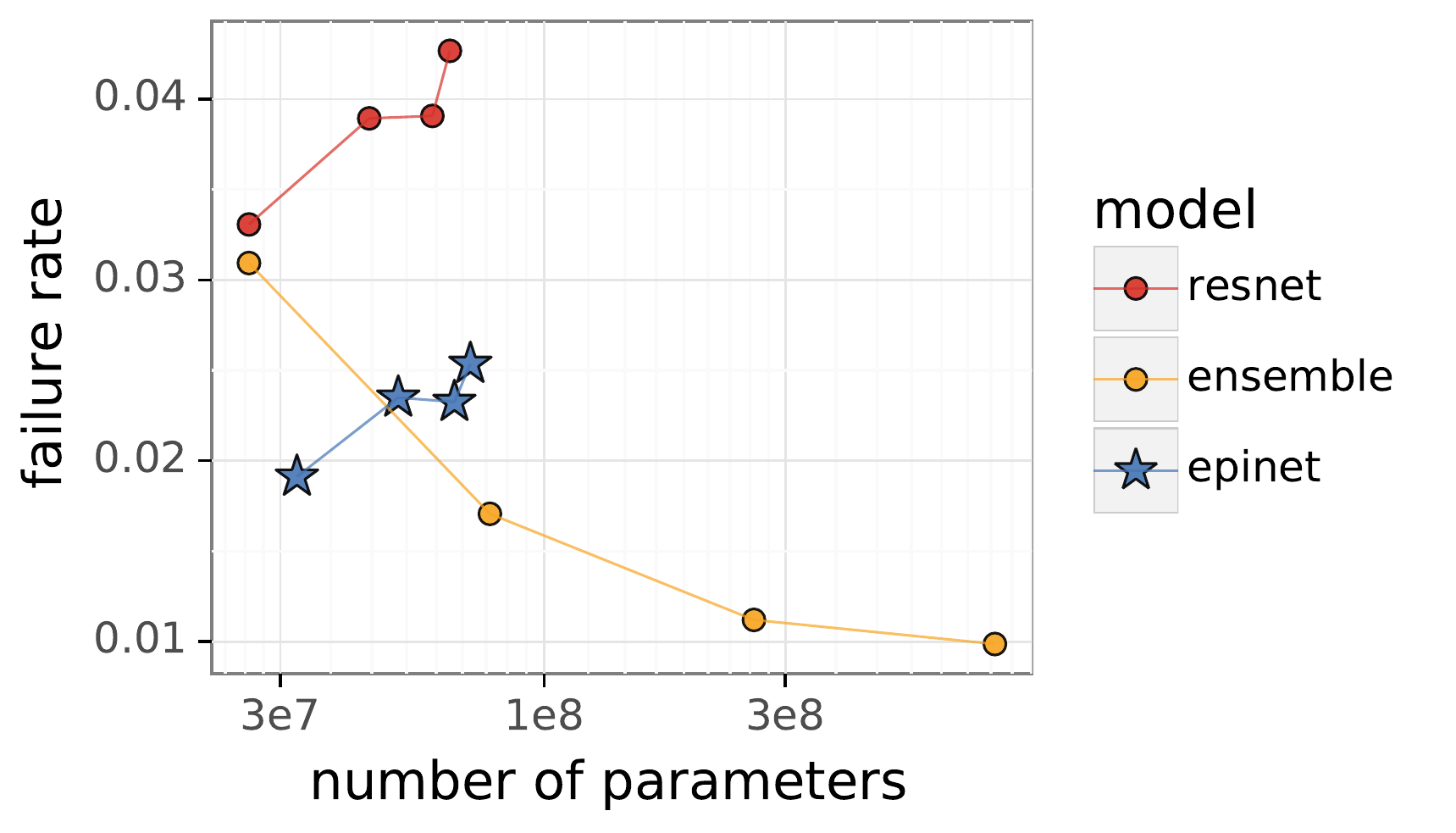}
    \caption{Percentage of ImageNet-A test examples for which the model predicts incorrectly but with over $95\%$ certainty.}
    \label{fig:imagenet_a_failure_rate}
\end{subfigure}
\caption{ImageNet-A prediction uncertainty and failure rate.}
\label{fig:imagenet_a_uncertainty}
\end{figure}

Figure~\ref{fig:imagenet_a_confidence} shows the benchmark models' confidence scores on ImageNet-A. Interestingly, the confidence scores are actually not very high, ranging from $34\%$ to $42\%$. Unlike the examples shown in \cite{hendrycks2021nae}, for which a ResNet-50 model makes the wrong predictions with $99\%$ certainty, here we see that all of our models are actually on average ambiguous about the correct classes.
In Figure~\ref{fig:imagenet_a_failure_rate}, we define the failure rate as the percentage of ImageNet-A examples for which the model makes the wrong prediction with over $95\%$ certainty. We see that the failure rate is below $5\%$ for all of the models. 
The epinet has a lower failure rate than the ResNet. 
Even though the failure rate of the ensemble is the lowest, recall from Figure~\ref{fig:imagenet_a} that the accuracy of the ensemble is a lot lower than the other two approaches.

\begin{figure}[t]
\centering
\includegraphics[width=0.7\textwidth]{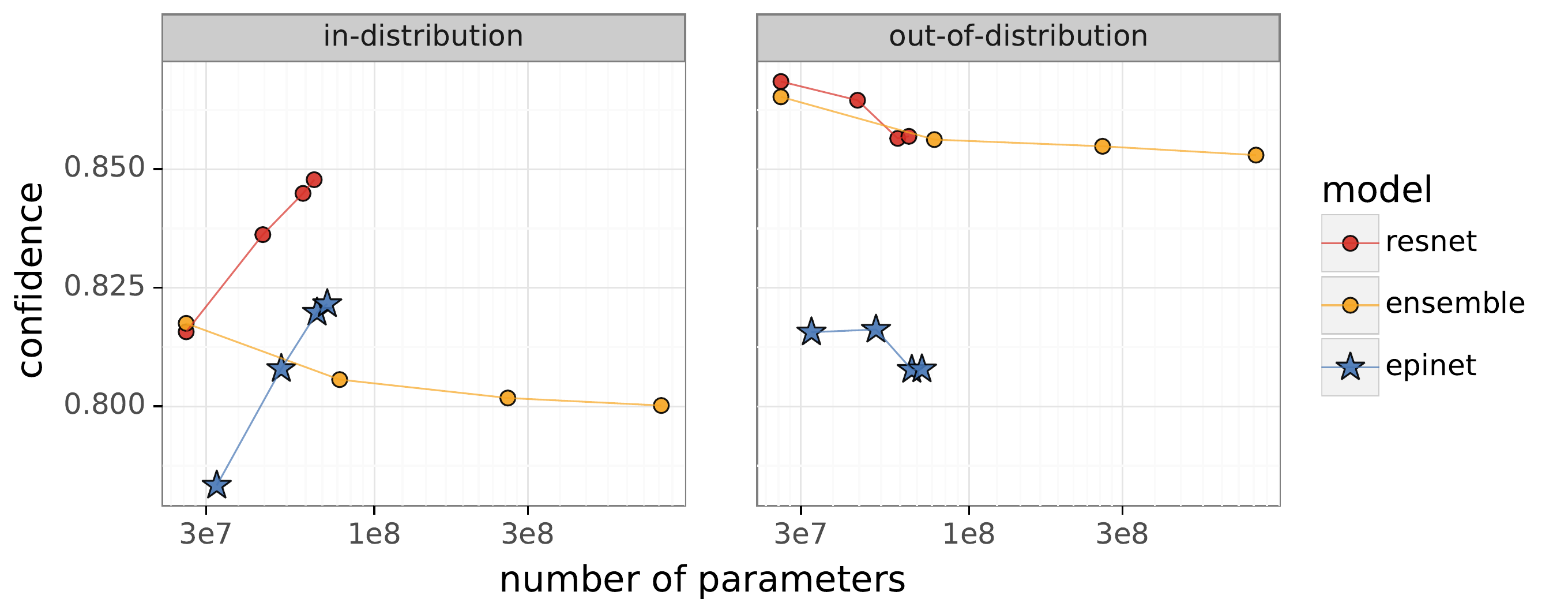}
\caption{Prediction confidence of benchmark models on ImageNet test examples (in-distribution) and ImageNet-O test examples (out-of-distribution).}
\label{fig:imagenet_o_uncertainty}
\end{figure}

Figure~\ref{fig:imagenet_o_uncertainty} shows the confidence scores of the benchmark models on the out-of-distribution ImageNet-O dataset and the in-distribution ImageNet dataset (restricted to the $200$ selected classes). Ideally, we would want the model to assign low confidence to out-of-distribution inputs and high confidence to in-distribution inputs. We see from Figure~\ref{fig:imagenet_o_uncertainty} that, unfortunately, all of the models fail to do this.
For the ResNet and ensemble models, the out-of-distribution confidence scores are in fact higher than the in-distribution scores.
The epinet seems to behave slightly better than the baselines, but still it can hardly distinguish out-of-distribution from in-distributuion inputs according to these confidence scores.
It seems disappointing that these models give such over-confident predictions on ImageNet-O, even though they seem to demonstrate reasonable levels of confidence on ImageNet-A. One hypothesis is that since the images in ImageNet-A actually belong to the classes present in the training set, these models might to some degree recognize features relevant to the correct classes, even though there might be other distracting features, leading to moderate confidence overall. In contrast, the images in ImageNet-O do not belong to any of the training classes, and, as such, it might be more likely that certain features dominate the predictions. This might be an interesting hypothesis to investigate in the future.
Another interesting observation is that, for the ResNet and epinet models, the in-distribution confidence scores increase with the model size, while the out-of-distribution confidence scores decrease with the model size. It appears that larger models improve generalization and robustness in this case.

\begin{figure}
\centering
\includegraphics[width=0.45\textwidth]{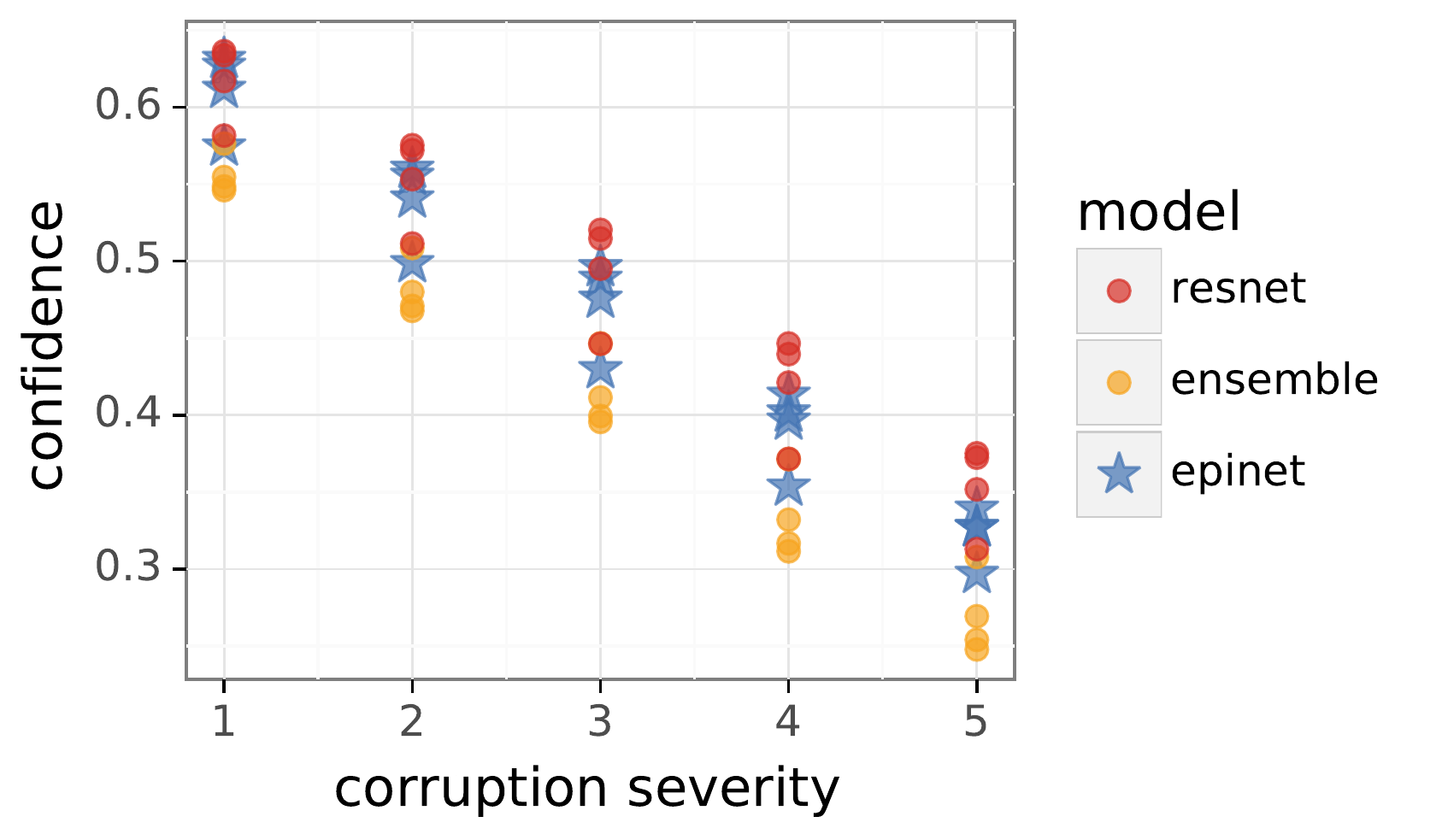}
\caption{Confidence scores of benchmark models as a function of corruption severity on ImageNet-C.}
\label{fig:imagenet_c_uncertainty}
\end{figure}

Figure~\ref{fig:imagenet_c_uncertainty} shows the average confidence scores of benchmark models on ImageNet-C as the level of corruptions applied to the test images becomes more severe. For each value of corruption severity, we average the confidence scores across all types of corruption noise. We see that all models appear reasonable in that, as the corruption noise becomes larger, the confidence scores become lower. It is interesting that none of these models are explicitly exposed to any of these corruption noises in their training, and yet they demonstrate a monotonic decrease in their prediction confidence as the corruption noise grows larger.

Overall, the benchmark models seem to demonstrate reasonable levels of prediction confidence on ImageNet-A and ImageNet-C, but all of them are extremely over-confident on ImageNet-O. More research is needed to investigate what kind of prior is needed to curb the over-confidence.

\subsection{Temperature re-scaling}
\label{se:temperature}
A common heuristic to improve model performance during evaluation is to re-scale the logits post training using a tunable ``temperature'' parameter. The heuristic is also applied to the ResNet, epinet, and ensemble models in \cite{osband2022epistemic}. In this section, we will look at these temperature re-scaled models that are improved on the ImageNet dataset and evaluate them on Imagenet-A/O/C.
Note that we do not consider having access to validation sets containing ImageNet-A/O/C samples, which poses a slightly different problem that deserves future investigation. Here we are interested in whether or not models equipped with temperature re-scaling optimized for the ImageNet dataset are robust on ImageNet-A/O/C.

\begin{figure}[t]
\centering
\begin{subfigure}{\textwidth}
\centering
\includegraphics[width=\textwidth]{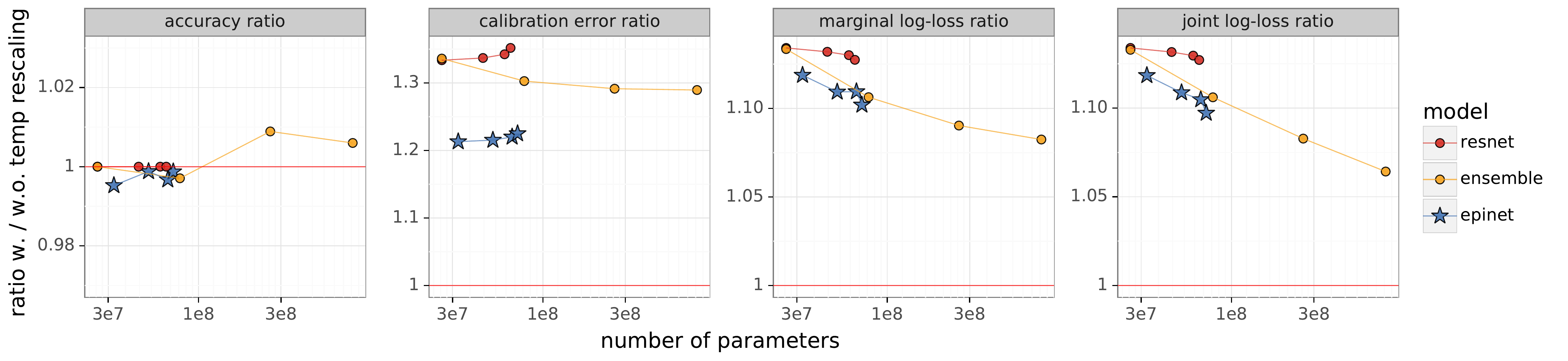}
\caption{ImageNet-A}
\label{fig:imagenet_a_temperature}
\end{subfigure}

\vspace{0.5cm}
\begin{subfigure}{\textwidth}
\centering
\includegraphics[width=0.4\textwidth]{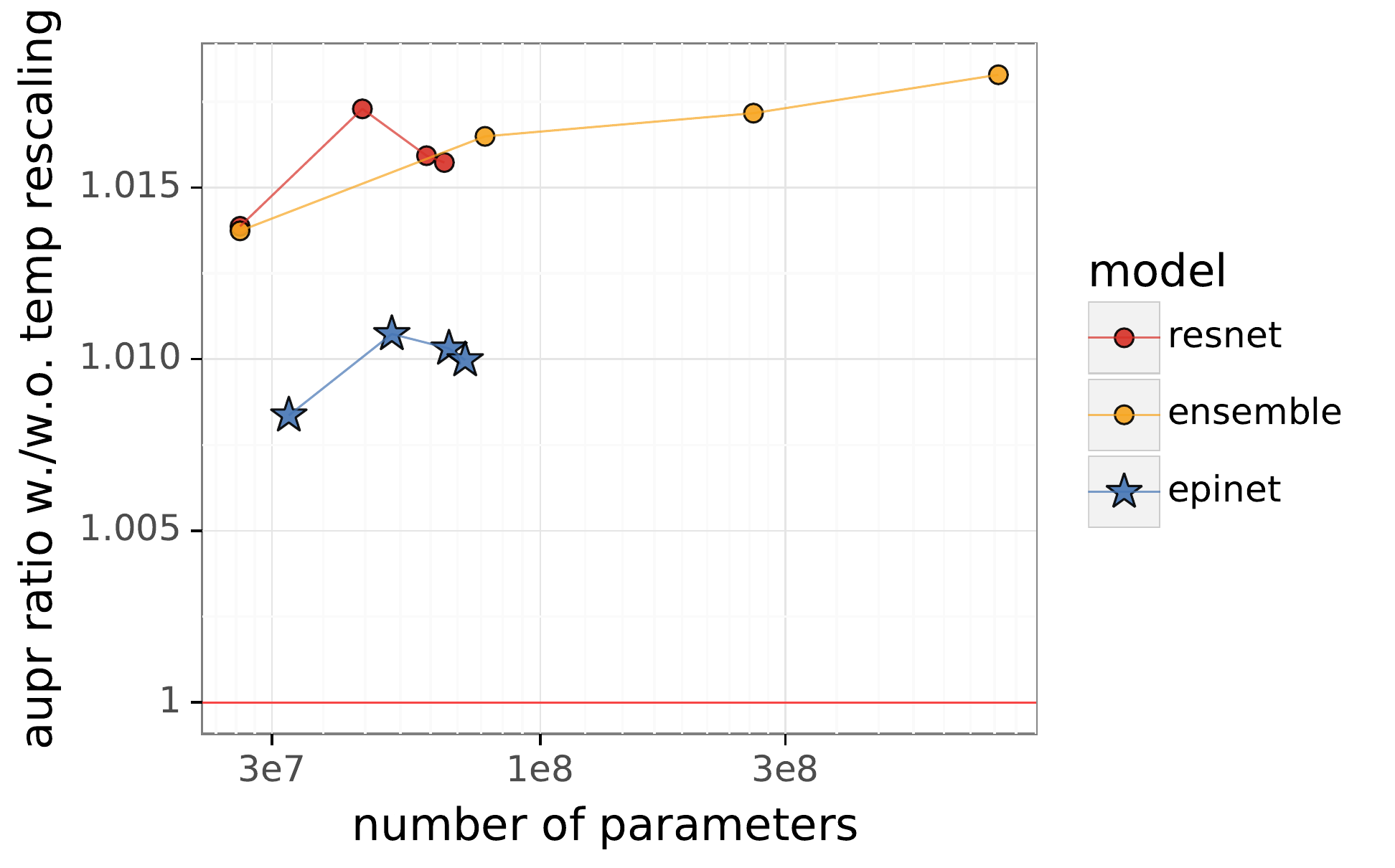}
\caption{ImageNet-O}
\label{fig:imagenet_o_temperature}
\end{subfigure}

\vspace{0.5cm}
\begin{subfigure}{\textwidth}
\centering
\includegraphics[width=\textwidth]{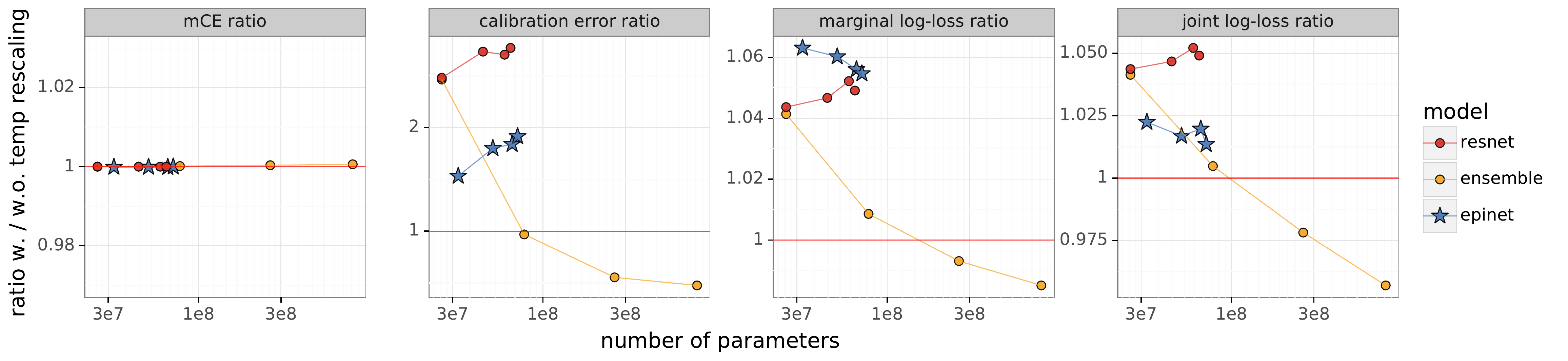}
\caption{ImageNet-C}
\label{fig:imagenet_c_temperature}
\end{subfigure}

\caption{Compare model performance on ImageNet-A/O/C with and without logits re-scaling using the temperatures optimized for the ImageNet evaluation set.}
\label{fig:temperature}
\end{figure}

In Figure~\ref{fig:temperature}, we show the ratio of model performance with and without temperature re-scaling on ImageNet-A/O/C. For the expected calibration error, marginal log-loss, and joint log-loss, a ratio above $1$ means that re-scaling by the temperature tuned for ImageNet evaluation hurts performance on the corresponding test set. For AUPR, a ratio above $1$ indicates that the performance improves with temperature re-scaling. Temperature re-scaling should not affect accuracy and mCE, for which we should observe a ratio close to $1$.

We see in Figure~\ref{fig:imagenet_a_temperature} that re-scaling by temperatures optimized for ImageNet hurts the performance of all models on ImageNet-A.
Figure~\ref{fig:imagenet_o_temperature} shows that AUPR improves slightly with temperature re-scaling, but the increase is marginal (less than $2\%$ difference).
Figure~\ref{fig:imagenet_c_temperature} shows that temperature re-scaling worsens the performance of the ResNet and epinet on ImageNet-C. However, for the ensemble models, temperature re-scaling helps ensembles of size greater than $3$. According to \cite{osband2022epistemic}, the tuned temperature for the ensemble models is less than $1$. This seems consistent with the observations made in \cite{rahaman2021ensemble} that the calibration error of ensembles is sensitive to temperature re-scaling. The marginal and joint log-losses are also slightly better with temperature re-scaling for ensembles of size more than $3$, but the differences are small. 

Overall, it seems that re-scaling the trained models by temperatures optimized for the ImageNet dataset in general does not offer much benefit on these robustness datasets, and, in several cases, it even hurts model robustness by a significant amount.

\section{Conclusions}
We investigated the robustness of epinets against distributional shifts by taking the epinet model trained on ImageNet \citep{osband2022epistemic} and evaluating it on ImageNet-A, ImageNet-O, and ImageNet-C. We compared its performance with the performances of ResNets, which the epinet uses as its base network, and ensembles of ResNets. We found that the epinet in general improves or attains a similar level of robustness as its base ResNet according to traditional robustness metrics. The ensemble approach is not competitive with the epinet according to these metrics. For joint predictions, the epinet significantly outperforms the alternatives by a huge margin.

Despite these improvements attained by the epinet, it is still far from addressing the robustness challenges imposed by these datasets.
One possible future research direction is to investigate what kinds of prior can effectively inform the epinet, or other ENNs, of the possibility of inputs from a shifted distribution. The benchmark models in this paper all started from a relatively uninformed prior, which we have seen does not fare well in these robustness tasks. 
One example of producing a stronger prior effect is through data augmentation, which is a common heuristic in the robustness literature. The ENN framework can easily incorporate this technique through a change in the training loss function. Compared to traditional models that apply data augmentation, the ENN approach can offer the additional benefit of improving joint predictions across multiple inputs.
Another possibility for epinets is to regularize through augmentation in the feature space rather than in the input image space.
More research is needed to design and understand different kinds of priors in order to improve model robustness against distributional shifts.

\bibliographystyle{apalike}
\bibliography{references}

\end{document}